\documentclass[journal]{IEEEtran}
\usepackage{amsfonts,amssymb}
\usepackage{amsmath}
\usepackage{amsthm}

\usepackage{graphicx}
\usepackage{enumerate}
\usepackage{multirow}
\usepackage{longtable}
\usepackage{rotating}
\usepackage{calc}
\usepackage{subfig}
\usepackage{graphicx}
\usepackage{epstopdf}
\usepackage{graphics}
\usepackage{epsfig}
\usepackage{psfrag}
\usepackage{cases}

\usepackage{array,color}
\usepackage{tabularx}
\usepackage{multirow}
\usepackage{booktabs}
\usepackage{makecell}

\usepackage{algorithm}

\usepackage{algorithmic}

\usepackage{multirow}

\usepackage[numbers,sort&compress]{natbib}
\usepackage{url}
\usepackage{bm}
\usepackage{amsfonts}
\usepackage{amsmath}
\usepackage{amssymb}
\usepackage{amsthm}
\usepackage{color}

\usepackage{array,color}%
\usepackage{tabularx}%
\usepackage{multirow}%
\usepackage{booktabs}%
\usepackage{makecell}  

\usepackage{graphicx}  
\usepackage{subfig}
\usepackage{epstopdf}
\usepackage{graphics}
\usepackage{epsfig}
\usepackage{psfrag}
\usepackage{overpic}

\usepackage{stfloats}
\usepackage{float}

\usepackage{indentfirst} %
\usepackage{gensymb}

\begin{document}

\title{An Accurate  and Real-time Self-blast Glass Insulator Location Method Based On Faster R-CNN and U-net with Aerial Images}
\author{Zenan Ling$^2$, Robert C. Qiu$^{1,2}$,~\IEEEmembership{Fellow,~IEEE}, Zhijian Jin$^2$,~\IEEEmembership{Member,~IEEE}

Yuhang Zhang$^2$, Xing He$^2$, Haichun Liu$^2$£¬ Chu Lei$^2$
\thanks{This work was partly supported by N.S.F. of  China  No.61571296 and N.S.F.  of  US  Grant No.  CNS-1247778, No.  CNS-1619250.

$^1$ Electrical and Computer Engineering Department,
Tennessee Technological University, Cookeville, TN 38505 USA. Dr. Qiu is also with Department of Electrical Engineering,
Research Center for Big Data Engineering Technology, State Energy Smart Grid Resarch and Development Center, Shanghai Jiaotong
 University, Shanghai 200240, China. (e-mail: rcqiu@sjtu.edu.cn)

$^2$ Department of Electrical Engineering,
Research Center for Big Data Engineering Technology, State Energy Smart Grid Resarch and Development Center, Shanghai Jiaotong
 University, Shanghai 200240, China. (e-mail: ling\_zenan@163.com; hexing\_hx@126.com; leochu@sjtu.edu.cn)
}}

\maketitle
\IEEEpeerreviewmaketitle

\begin{abstract}
The location of broken insulators in aerial images is a challenging task. This paper, focusing on the self-blast glass insulator, proposes a  deep learning solution.
We address the  broken insulators location problem as a low signal-noise-ratio image location framework  with two modules: 1) object detection based on Fast R-CNN, and 2) classification of pixels based on U-net. A diverse aerial image set of some grid in China is tested to validated the proposed approach. Furthermore, a comparison is made among different methods and the result shows that our approach is accurate and real-time.

Keywords- insulators; location; aerial images; deep learning; real-time; faster r-cnn; U-net
\end{abstract}
\section{INTRODUCTION}
\subsection{Background}
\label{background}
\begin{figure}[htbp]
 \centering
 \subfloat{\label{fig1a}
 \includegraphics[width=0.2\textwidth]{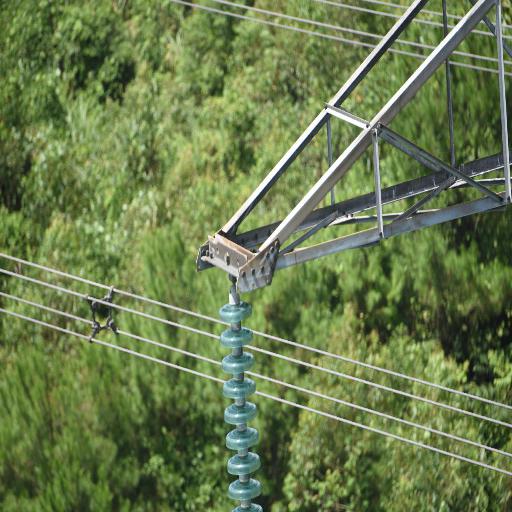}
 \label{fig1.a}}
 \subfloat{\label{fig1b}
 \includegraphics[width=0.2\textwidth]{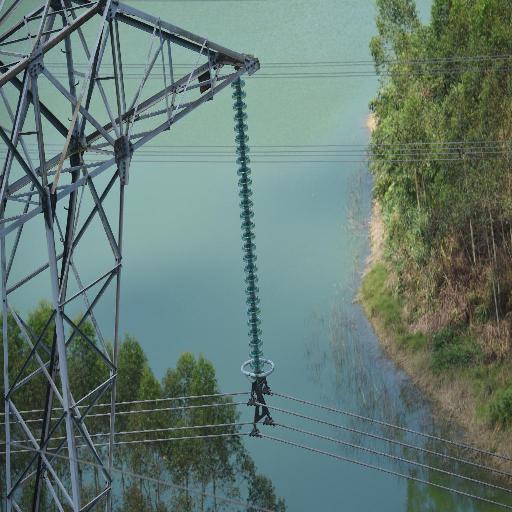}
 \label{fig1.b}}
 \\
 \subfloat{\label{fig1c}
 \includegraphics[width=0.2\textwidth]{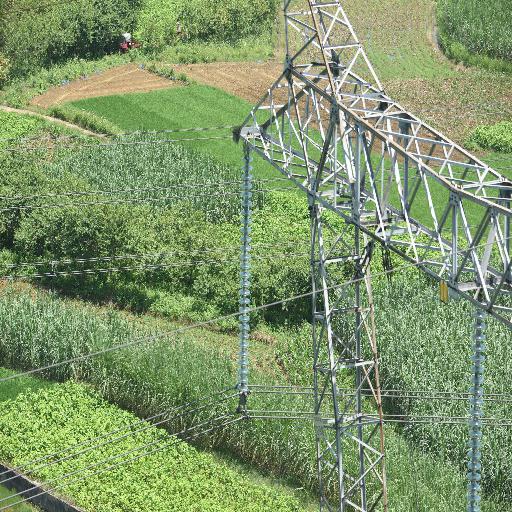}
 \label{fig1.c}}
 \subfloat{\label{fig1d}
 \includegraphics[width=0.2\textwidth]{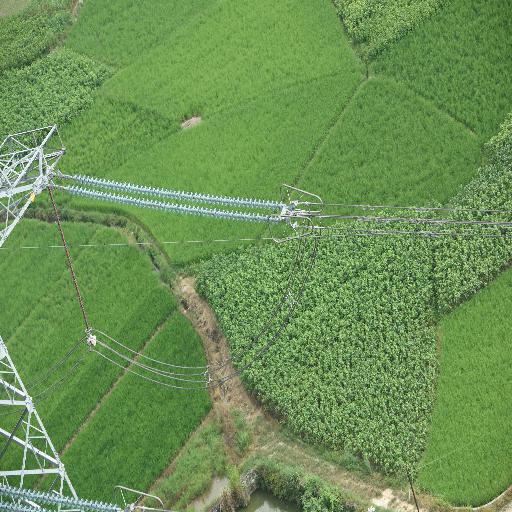}
 \label{fig1.d}}
 \caption{The aerial images captured by helicopters or UAV.}
\label{insulator_pic}
 \end{figure}

Insulators are widely used in high voltage transmission lines and play a significant role in the electrical insulation and conductor conjunction. The insulators fault, e.g.,  glass insulators self-blast, poses a grave threat to  power systems as it could cause  a cascade failure. It is of great safety risk to conduct periodical manual inception of insulators under extremely high voltage conditions, as a result that thousands of  insulators are deployed on the  transmission lines, which are usually long in distance and high in altitude.
The common approach is to capture aerial images (see Fig.~\ref{insulator_pic}) of insulators by manned helicopters or Unmanned Aerial Vehicles (UAV). As the humongous aerial images are more and more easier-accessible, an \emph{accurate} and \emph{real-time} broken insulators location method is in urgent need.

\begin{figure}[htbp]
 \centering
 \subfloat{\label{fig2a}
 \includegraphics[width=0.2\textwidth]{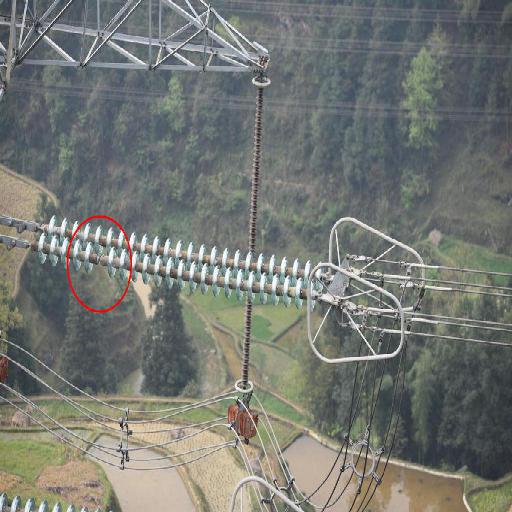}
 \label{fig2.a}}
 \subfloat{\label{fig2b}
 \includegraphics[width=0.2\textwidth]{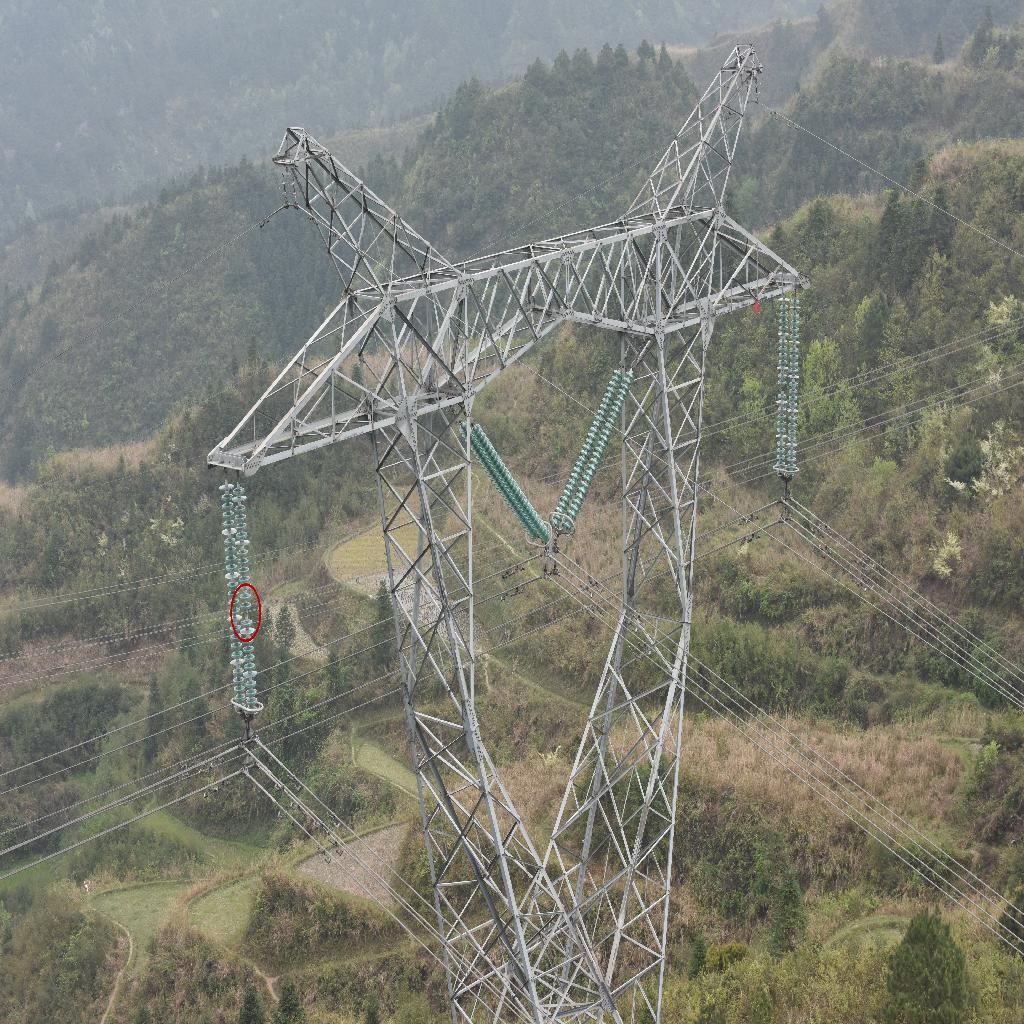}
 \label{fig2.b}}
 \caption{The  self-blast glass insulators: the broken parts are labelled by red circle.}
\label{badimage}
 \end{figure}

General speaking, following factors, as shown in  Fig.~\ref{badimage},  present a challenge to the broken insulator location problem:

\begin{enumerate}[1)]
\item Complicated background: the backgrounds of aerial images  often include various scenes such as forests, rivers, farmlands and so on.
\item Dynamic view changing: sometimes the camera may be not brought into focus, causing  non-distinctive feature even to human beings.
\item Low signal-noise-ratio (SNR): compared to the  whole  image,  quite a few pixels  contain the broken insulator information. In some extreme cases, the broken position can hardly be located  via human eye identification.
\end{enumerate}

In the last decade, deep learning  has achieved remarkable development especially in computer vision. AlexNet~\cite{Krizhevsky2012ImageNet}, proposed by Hinton et al., leads to a  new wave of deep learning.
Along with the invention of the new network architectures such as VGG~\cite{Simonyan2014Very}, InceptionNet~\cite{Szegedy2014Going}, ResNet~\cite{He2015Deep}, deep learning  beats the classical methods in various image \emph{classification} competitions. Furthermore, Faster R-CNN~\cite{Ren2015Faster}, proposed by Ren et al., realizes trained end-to-end and real-time \emph{object detection} with high accuracy. Besides, fully convolutional network (FCN), proposed by~\cite{Shelhamer2017Fully}, produces accurate and detailed  pixel-to-pixel segmentations and improves the previous best results in \emph{semantic segmentations}.

\subsection{Related Work}
Recently, much work has been done on the insulator detection and status classification. Zhao et al.~\cite{Zhao2016Multi} proposes a deep CNN model with multi-patch feature extraction   and a Support Vector Machine (SVM) for insulator status classification.  Liao et al.~\cite{Liao2017A} proposes a robust insulator detection algorithm based on local features and spatial orders for aerial images. This feature extraction method is model-based and able to locate the insulator string. Wu et al.~\cite{Wu2012A} uses the global minimization active contour model (GMAC) for insulator segmentation. Reddy et al.\cite{Reddy2013Condition} uses SVM and  Discrete Orthogonal S Transform (DOST) to carry out condition monitoring of  insulators in a complex background. This method needs to transform the raw image to complex frequency-domain which may cause extra computation complexity.

Little work, however, has been done about the  broken insulator location based on deep learning, which is the state of art architectures for computer vision.  Broken insulator location for aerial images is a typical computer vision problem.

\subsection{Contributions of Our Paper}
This paper focuses on the self-blast glass insulator location problem.
 The contributions are concluded as follows:
 \begin{enumerate}
\item The self-blast glass insulator location problem is formulated into  two computer vision problems: object detection and semantic segmentation.
\item Two state-of-art deep learning architectures are introduced: Faster R-CNN~\cite{Ren2015Faster} and U-net~\cite{Ronneberger2015U}. To our best knowledge, there is no similar work reported in this domain.
\item The object detection module realizes the insulator string location as  a by-product  of our method.
\item The proposed method is  \emph{accurate} and \emph{real-time}.
\end{enumerate}

The rest of this paper is organized as follows. Section~\ref{methodintro} proposes  the self-blast glass insulator location method.  Section~\ref{faster_rcnn} and~\ref{unet} briefly introduce the architectures of Faster R-CNN and U-net. Section~\ref{experiment} gives the location results in our experiment, and makes a comparison with different methods.
Section~\ref{conclusion} concludes the paper.
\section{ SELF-BLAST  GLASS INSULATOR  LOCATION METHOD}
\label{methodintro}

\subsection{Self-blast Glass Insulator Location Using Aerial Images}
The challenges in broken insulator location, mentioned in~\ref{background}, drive us to focus on following three aspects:

\subsubsection{The efficient feature extraction of the aerial images}
We choose CNN based methods because they are proved to preform much better than other methods in multiple image tasks. If the CNN model is trained directly, however, it is hard to  achieve desired results as the limitation of the sample numbers. Borrowing the idea of \emph{transfer learning}, we adopt the convolutional layers pre-trained on large-scaled image libraries.
\subsubsection{The enhancement of SNR}
It is natural to crop  the parts containing the insulator strings to improve SNR. Thus, the problem is transformed into  the location of insulator strings. Faster R-CNN is suitable to this task. It has the \emph{state-of-art} performance in object detection or location tasks; moreover, the pre-trained convolutional layers are adopted in its architecture (see Section~\ref{pretrainend}).
\subsubsection{The judgement of insulators' states}
Based on the above analysis, the cropped images may have different sizes because of  the sizes of the insulator strings, as well as their shooting angles, are different.
Standard CNN architectures are not able to processing images of arbitrary sizes. As a result, U-net, a specific FCN, is introduced to handle this problem. Besides, U-net is widely used in semantic segmentation tasks especially  in  biomedical image segmentation~\cite{Ronneberger2015U}, which is often  a low SNR task. Taking account of above mentioned reasons, U-net is adopted as a part of our method to conduct the binary classification of the pixels  in the  aerial images.

\begin{figure*}[htpb]
\centering
\begin{overpic}[scale=0.5
]{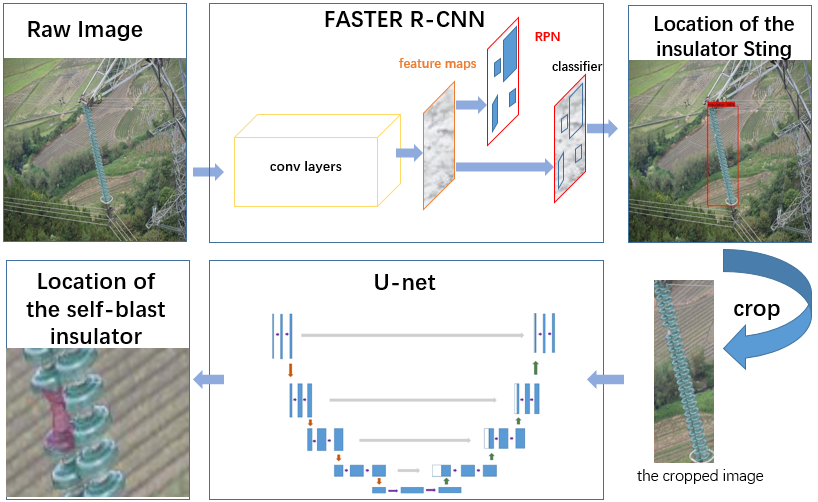}
\end{overpic}
\caption{Overall flow chart of the proposed method}
\label{fig:procedure}
\end{figure*}

Based on the above analysis, the self-blast glass insulator location method is designed.  The proposed method contains two modules: 1) the object detection module based on Faster R-CNN, and 2) the segmentation module based on U-net. The procedure is shown as Fig.~\ref{fig:procedure}.

Firstly, Faster R-CNN is adopted to locate the insulator stings. The insulator strings, in the form of rectangular  boxes, are cropped from the raw image, and the other parts are dropped. In this way, the  redundant information in the aerial images are thrown away to enhance SNR.

Secondly, take the cropped images as the input of  U-net, and the pixel binary classification results as the output---for broken parts, the  pixels are labelled as false and for normal parts as true.

The final output of the proposed method contains two parts: the locations of the insulator strings, and the  coordinates of  the broken  insulators.

\subsection{Real-time Performance of Proposed Method}
On the one hand, Faster R-CNN enables a unified, deep-learning-based object detection system to run at near real-time frame rates through making  region proposal network (RPN) sharing convolutional features with the down-stream detection network.

On the other hand, there is no fully connected layer contained in the architecture of U-net so that the number of parameters are  reduced dramatically. In practice, the classification of a single image costs extremely little time based on U-net.

Besides, the proposed method can be easily expanded into a  \emph{distributed system} by using Graphics Processing Units (GPUs). Multiple images can be processed simultaneously provided that sufficient GPUs are available.

For the integrity of this paper, a brief introduction of Faster R-CNN and U-net is given in the next two sections.
\section{INSULATOR STRING LOCATION METHOD: FASTER R-CNN}
\label{faster_rcnn}
\subsection{Structure of Faster R-CNN}
\begin{figure}[htpb]
\centering
\begin{overpic}[scale=0.28
]{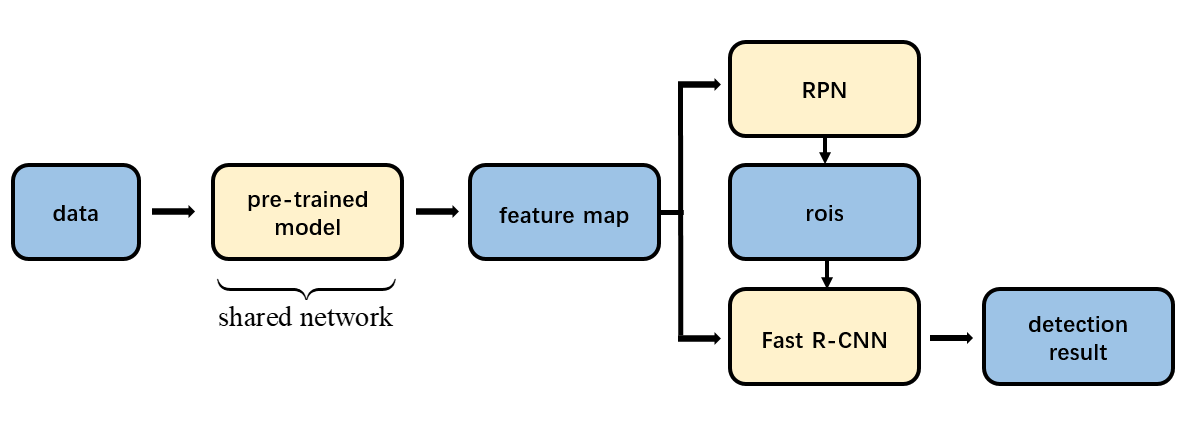}
\end{overpic}
\caption{The network structure of Faster R-CNN}
\label{fig:Faster R-CNN}
\end{figure}

Faster R-CNN is composed of two networks: RPN and Fast R-CNN. RPN is a fully convolution network for generating the proposal regions, and Fast R-CNN conducts detection and classification based on the proposal regions. The architecture of Faster R-CNN is illustrated in Fig.~\ref{fig:Faster R-CNN}.

\subsubsection{Feature Extraction}
\label{pretrainend}
As shown in Fig.\ref{fig:Faster R-CNN}, efficient and rich features of the aerial images  are extracted by a  commonly used CNN model  pre-trained on ImageNet such as Resnet, VGG.  The pre-trained convolutional layers are shared by RPN and Fast R-CNN.
\subsubsection{RPN}
\begin{figure}[htpb]
\centering
\begin{overpic}[scale=0.5
]{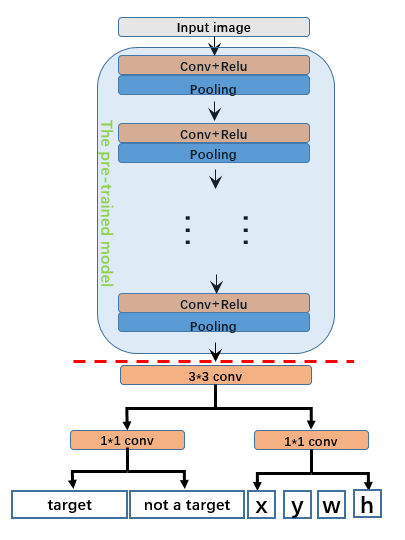}
\end{overpic}
\caption{The network structure of RPN}
\label{fig:RPN}
\end{figure}

RPN is the core part of Faster R-CNN. As shown in Fig.~\ref{fig:RPN}. RPN aims to identify all possible candidate boxes.  3*3 convolution kernels connected to the last shared convolutional layer are used to generate a feature map of 256 channels. Sliding window method is conducted over the feature map, and a 256 dimension feature vector is obtained at each location of the feature map. Within each sliding-window, the region proposals are predicted simultaneously. The proposals are parameterized relative to reference boxes which are called "anchors".  Nine anchors, composed of 3 different sizes and 3 different scales, are adopted to improve the accuracy of proposal regions. The candidate boxes are relatively sparse due to the subsequent location refinement procedure.
\subsubsection{Fast R-CNN}
The proposal region generated from RPN is detected by Fast-RCNN network. Reshape the feature map as a high dimensional feature vector. This feature layer is fully connected with  another feature layer of the same length. The  following parameters are predicted: the probability value of that the candidate box belongs to  each class; the parameters for the bounding box's translation and scaling.

\section{SELF-BLAST INSULATOR LOCATION METHOD:  U-NET}
\label{unet}
U-net is a kind of  FCN architecture. There are two core operations contained in FCN: 1) \emph{up sampling} and 2) \emph{skip layers}. FCN uses up sampling to ensure that the output has the same size as the input image. However,  the feature map obtained by direct up sampling often leads to very rough segmentation results. Thus, another structure \emph{skip layers} is adopted to overcome this problem. Skip layers method  conducts \emph{up sampling} for the  outputs of different pooling layers, and refines the  results by feature fusion such as copying and cropping.
\subsection{The Structure of U-net}
\begin{figure}[htpb]
\centering
\begin{overpic}[scale=0.29
]{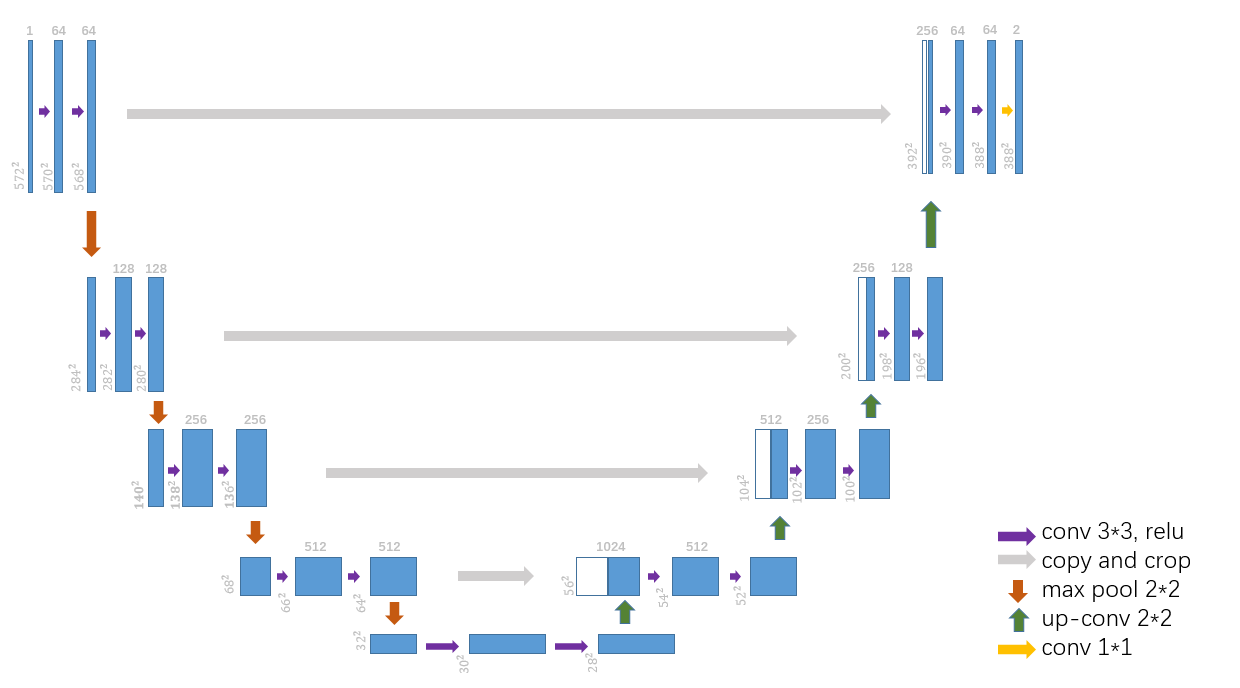}
\end{overpic}
\caption{The network structure of U-net}
\label{fig:U-net}
\end{figure}
The structure shown in Fig.\ref{fig:U-net} and its shape is like U. Note that the input of U-net can be an image of arbitrary size, e.g.  572*572 in Fig.~\ref{fig:U-net}.

U-net is composed of the compression and recovery layers. The compression layers is a classical VGGNet-16~\cite{Simonyan2014Very}. The structure of two 3*3 convolution kernels and a 2*2 max-pooling layer are adopted repeatedly and  more high-level features are extracted layer-by-layer using the down sampling effect of the pooling layer. The structure of the recovery layers is totally different from that of VGGNet-16. The structure of a 2*2 deconvolution kernel and two 3*3 convolution kernels are adopted repeatedly.

Finally,  by a 1*1 convolutional kernel, the dimension of the feature map is mapped to 2. And softmax is used for classification in the output layer.

A remarkable detail is that the input of the pooling layer must be a feature map with even height and width. Thus the hyper-parameter needs to be set carefully.

\section{Experiment}
\label{experiment}

In this section, we present our results on the aerial images provided by China South Grid. Firstly, the data set, computation source, the preprocessing method and the construction of training  and test sets are  introduced.  Secondly,   details of the training and testing of two network architecture are reported. The accuracy and speed of the proposed method is evaluated and  compared with other methods. Parts of the results of the proposed method are presented by images. Besides, the advantage of the combination of the two models and   the influence of the number of samples are also discussed.
\subsection{Experiment Description }
\subsubsection{Data and Computation Source}
\begin{figure}[htbp]
 \centering
 \subfloat[a positive sample]{\label{samplep}\includegraphics[width=0.3\textwidth]{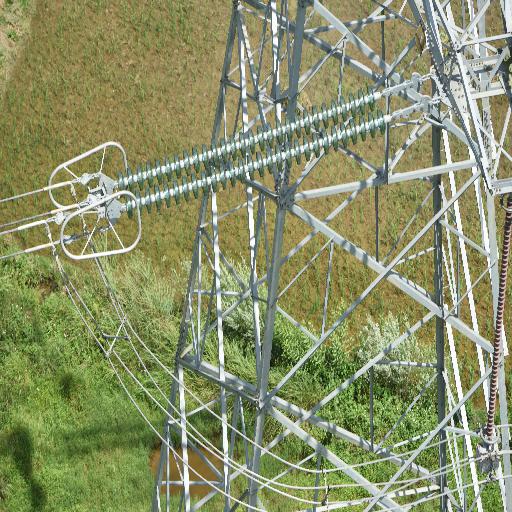}}\\
 \subfloat[a negative sample]{\label{samplen}\includegraphics[width=0.3\textwidth]{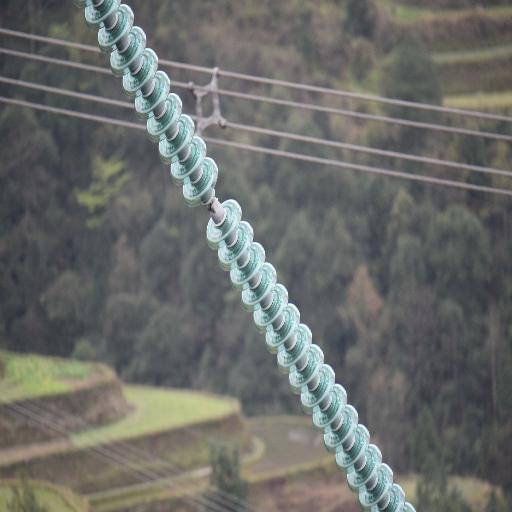}}
 \caption{The examples of training samples}
\label{examples}
 \end{figure}

The experiment data set is 620 high  resolution images collected  by helicopters. The data set consists of 400 positive samples and 220 negative samples (see Fig.~\ref{examples}).   The size of raw images are 7360*4912 or 4512*3008. All the experiments are conducted on a computer with Intel Core i7-7700 (3.60GHz), 32GB RAM and two NVIDIA GTX 1080 with GPU memory of 8 GB.  Our implementation uses TensorFlow 1.2.0~\cite{tensorflow2015-whitepaper}.

\subsubsection{ Data Preprocessing}
Due to the limitation of computation resource, the raw images are resized to 1024*1024. Note that no other preprocessing operation like intercepting target subjects which is often adopted by other methods is conducted in our experiment.
\subsubsection{Construction of Training Sets and Test Sets}
Due to the number of sample is not large enough, 3-fold cross-validation i.e. 412 images for training and 208 images for testing is carried out in the experiment for insulator strings location. In the experiment for classification, the 4-fold cross-validation is conducted on the cropped images.  The insulator string is labelled with "insulator" by a box which is determined by four coordinates. Note that only glass insulators in the images are labelled. The pixels that belong to broken insulators are labelled with "break" and the others are labelled with "normal".
\subsubsection{Evaluation Method}
The performance of the proposed method is evaluated by the values of recall and precision defined as follows:
\begin{equation}\label{Precision}
  Precision = TP/(TP+FP)
\end{equation}
\begin{equation}\label{Recall}
  Recall = TP/(TP+FN)
\end{equation}
where $TP$ is the number of successfully located targets. $TP+FP$ is the total number of located objects, and $TP+FN$ is the total number of actual targets.
\subsection{The Insulator String Location}
\begin{figure}[htpb]
\centering
\begin{overpic}[scale=0.3]{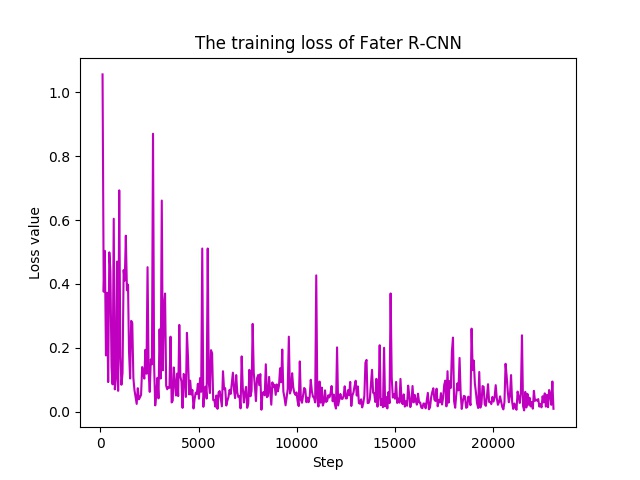}
\end{overpic}
\caption{The training loss of Faster R-CNN}
\label{fig:location_loss}
\end{figure}

The location of glass insulator strings is conducted through Faster R-CNN. The best  performance of Faster R-CNN is obtained with the pre-trained model inception-resnet-v2~\cite{Szegedy2016Inception}. The training of Faster R-CNN is carried out using Momentum method~\cite{Sutskever2013On}. The parameter settings are shown in Tab.~\ref{tab:parameter} and the training loss is plotted in Fig.~\ref{fig:location_loss}.

\begin{table}
  \centering
  \begin{tabular}{p{3.4cm} | p{0.6cm}}
\hline
  Parameter name & Value\\

\hline
  batch size & 2\\
  \
  max step & 23000\\

  dropout keep probability & 1.0\\
  the score threshold for NMS & 0.0\\
  the IoU threshold for NMS & 0.7\\
  the initial learning rate & 0.0003\\
  the momentum value & 0.9\\
\hline
\end{tabular}
  \caption{The parameter settings of Faster R-CNN}\label{tab:parameter}
\end{table}

\begin{figure}[htbp]
 \centering
 \subfloat{\label{fig8a}
 \includegraphics[width=0.2\textwidth]{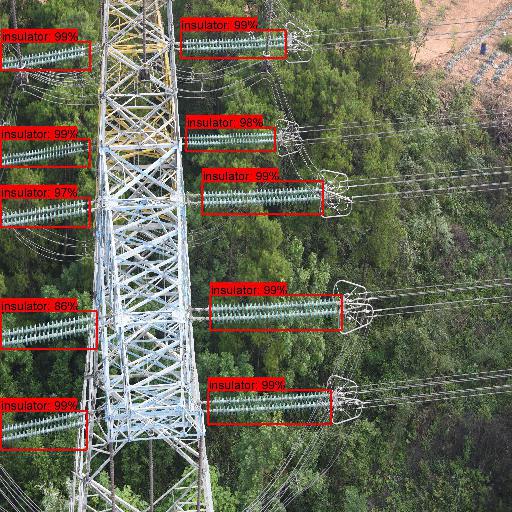}
 }
  \subfloat{\label{fig8b}
 \includegraphics[width=0.2\textwidth]{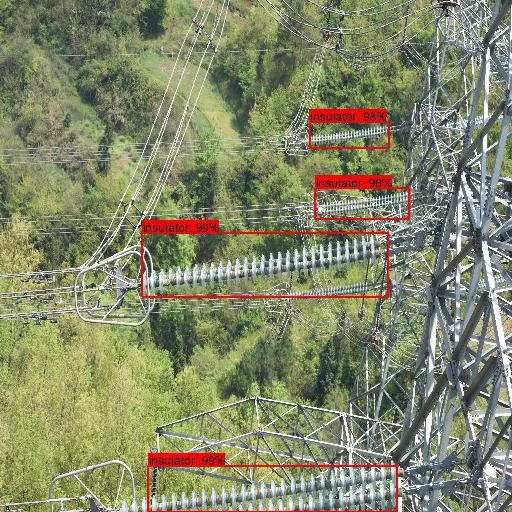}
 }\\
 \subfloat{\label{fig8c}
 \includegraphics[width=0.2\textwidth]{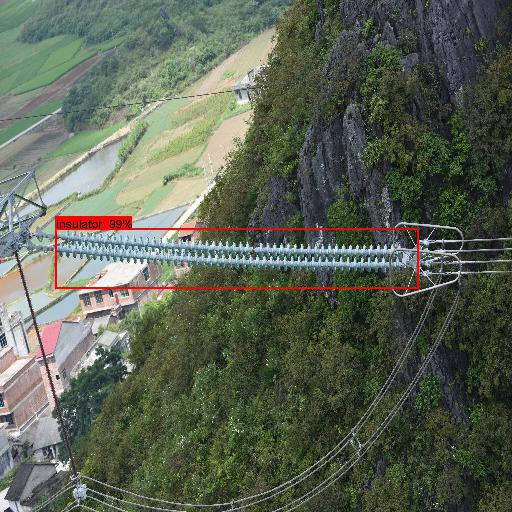}
 }
  \subfloat{\label{fig8d}
 \includegraphics[width=0.2\textwidth]{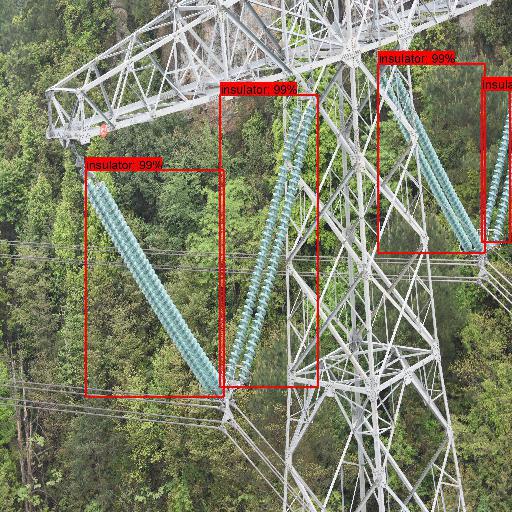}
 }
 \caption{Four Insulator String Location Results. Each detected insulator string is bounded by a red box.  Each output box is associated with a category label and a softmax score in $[0,1]$.}
\label{location}
 \end{figure}

Fig.~\ref{location} shows two location results. The detected insulator strings are bounded by red boxes and the scores and class name are shown above these boxes. As shown in Fig.~\ref{location}, the location method is fairly effective in detecting glass insulator strings under a complex scene.
If the overlapping portion of the result and the ground truth is greater than 0.95, we regard that the detection is successful. The value of precision is \textbf{0.966} and the value of recall  is \textbf{0.971}. Besides, the average time cost on a single image is \textbf{793 ms}. These results suggest that the method enables the real-time location of insulator strings with high accuracy.

We also test other three widely used object detection architectures on our data set: single shot multibox detector (SSD)~\cite{Liu2015SSD}, region-based fully convolutional networks  (RFCN)~\cite{Dai2016R} and discriminatively trained part-based models (DPM)~\cite{Szegedy2013Deep}. The comparison  of different detection architectures with different pre-trained CNN models is presented in Tab.~\ref{tab:premodel}.

\begin{table}
  \centering
  \begin{tabular}{p{1cm}|p{2.2cm}|p{1cm}|p{1cm}|p{1cm}}
\hline
  Model name & Pre-trained model & Precision & Recall & Speed (ms) \\

\hline
               & inception-resnet-v2 & \textbf{0.966} & \textbf{0.971} & 793 \\
 Faster  & resnet-101 & 0.96 & 0.96 & 236\\
      R-CNN    & resnet-50 & 0.948 & 0.942 & 120\\
               & inception-v2 & 0.947 & 0.931 & 69\\
            & vgg-16 & 0.916 & 0.922 & 25\\
\hline
  & inception-resnet-v2 & 0.953 & 0.942 & 403\\
  & resnet-101 & 0.947 & 0.925 & 111\\
  SSD & resnet-50 & 0.941 & 0.914 & 75\\
  & inception-v2 & 0.941 & 0.908 & 42\\
  & vgg-16 & 0.882 & 0.857 & 15\\
 \hline
  & inception-resnet-v2 & 0.953 & 0.942 & 467\\
   & resnet-101 & 0.947 & 0.941 & 178\\
  RFCN & resnet-50 & 0.931 & 0.936 & 96\\
    & inception-v2 & 0.924 & 0.910 & 60\\
    & vgg-16 & 0.909 & 0.857 & 19\\
 \hline
 DPM & & 0.771 & 0.731 & 900\\
\hline
\end{tabular}
  \caption{Performance of different object detection models}\label{tab:premodel}
\end{table}

 General speaking, DPM without the pre-trained model has a worst performance. The \emph{trade-off} between speed and accuracy is obvious. Comparing to Faster R-CNN, SSD and RFCN are faster in calculation and lower in accuracy. For the reason that the time cost are more tolerable (793 ms), we choose Faster R-CNN with  inception-resnet-v2 to purchase a better accuracy.

\subsection{Self-Blast Insulator Location Results}
\label{Classification Results}
\begin{figure}[htpb]
\centering
\begin{overpic}[scale=0.3]{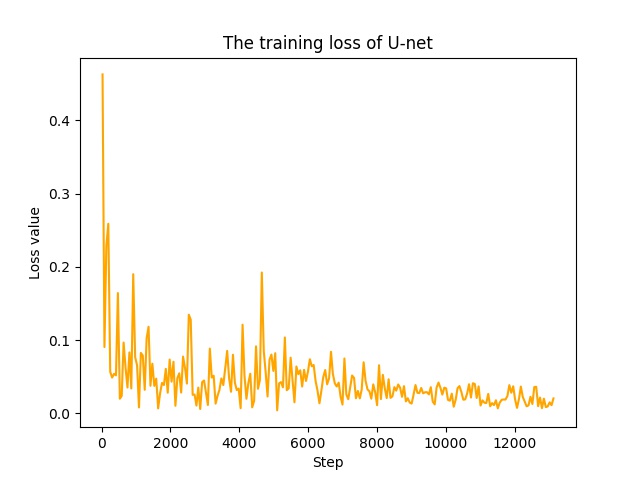}
\end{overpic}
\caption{The training loss of U-net}
\label{fig:classification_loss}
\end{figure}

Firstly, the images are cropped according to the bounding box as the input of U-net. The architecture used in our experiment is shown as  Fig.~\ref{fig:U-net}. Note that \emph{batch normalization} layers  are added after the convolutional layers to avoid  vanishing gradient problem and to accelerate deep network training~\cite{Sergey2015Batch}. Meanwhile, neither the drop out nor  weight regularization is adopted.  The optimization algorithm adopted in this experiment is Adam~\cite{Kingma2014Adam}. In Adam, the parameters $\epsilon$, $\beta_1$, $\beta_2$ are set to 1e-8, 0.9, 0.999 respectively (as recommended in~\cite{Kingma2014Adam}). The batch size is set to 1 and the initial learning rate is set to 1e-3. The data augmentation methods such as rotation and flip are carried out.  As shown in Fig.~\ref{fig:classification_loss}, the training loss converges after about 12000 steps.

\begin{figure*}[htbp]
 \centering
 \subfloat{\label{fig9a}\includegraphics[width=0.2\textwidth]{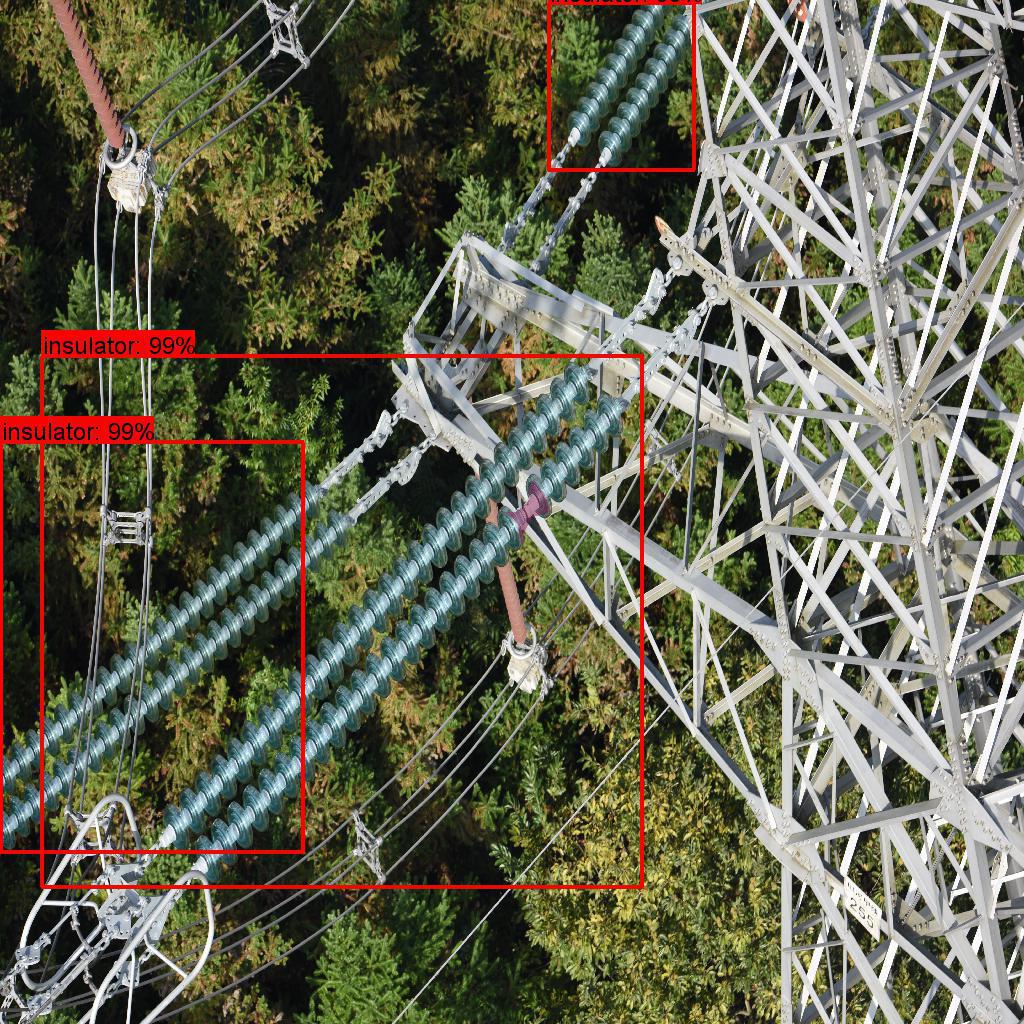}}
 \subfloat{\label{fig9b}\includegraphics[width=0.2\textwidth]{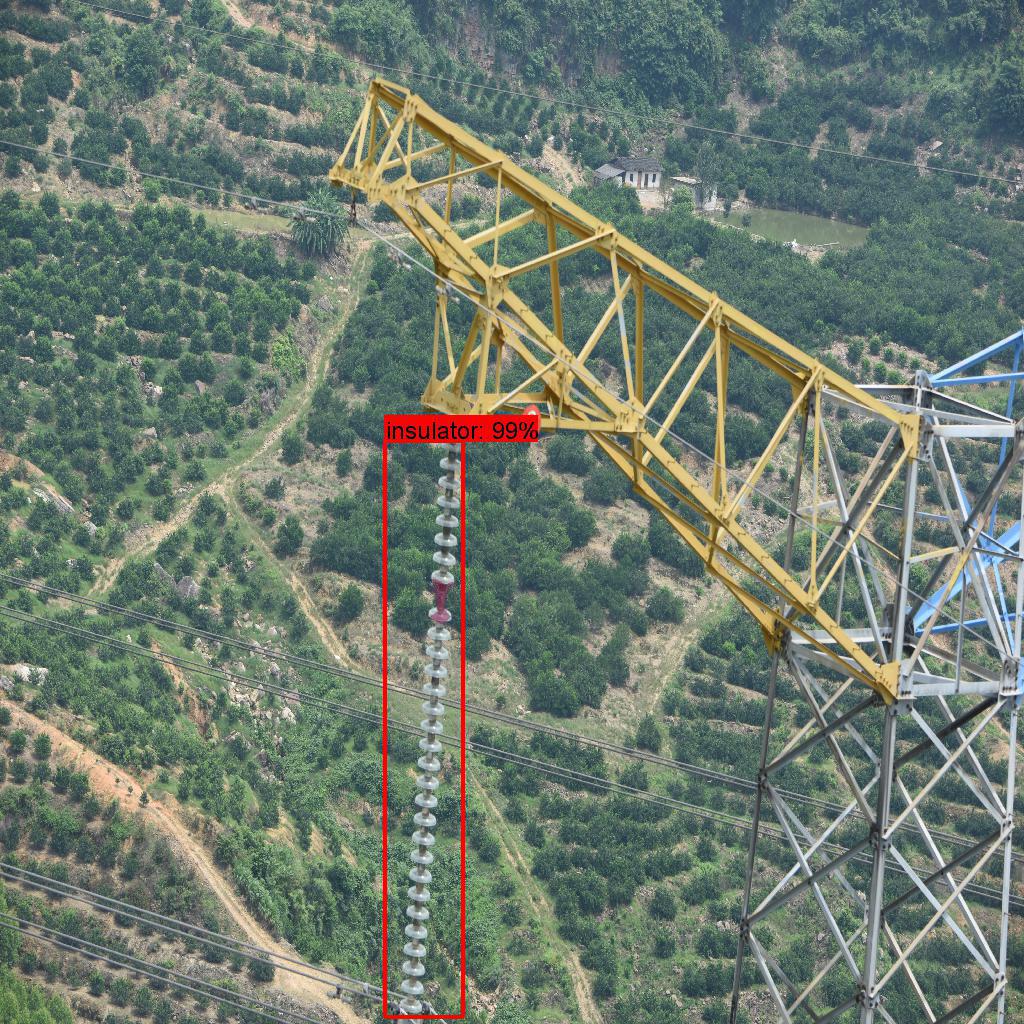}}
 \subfloat{\label{fig9d}\includegraphics[width=0.2\textwidth]{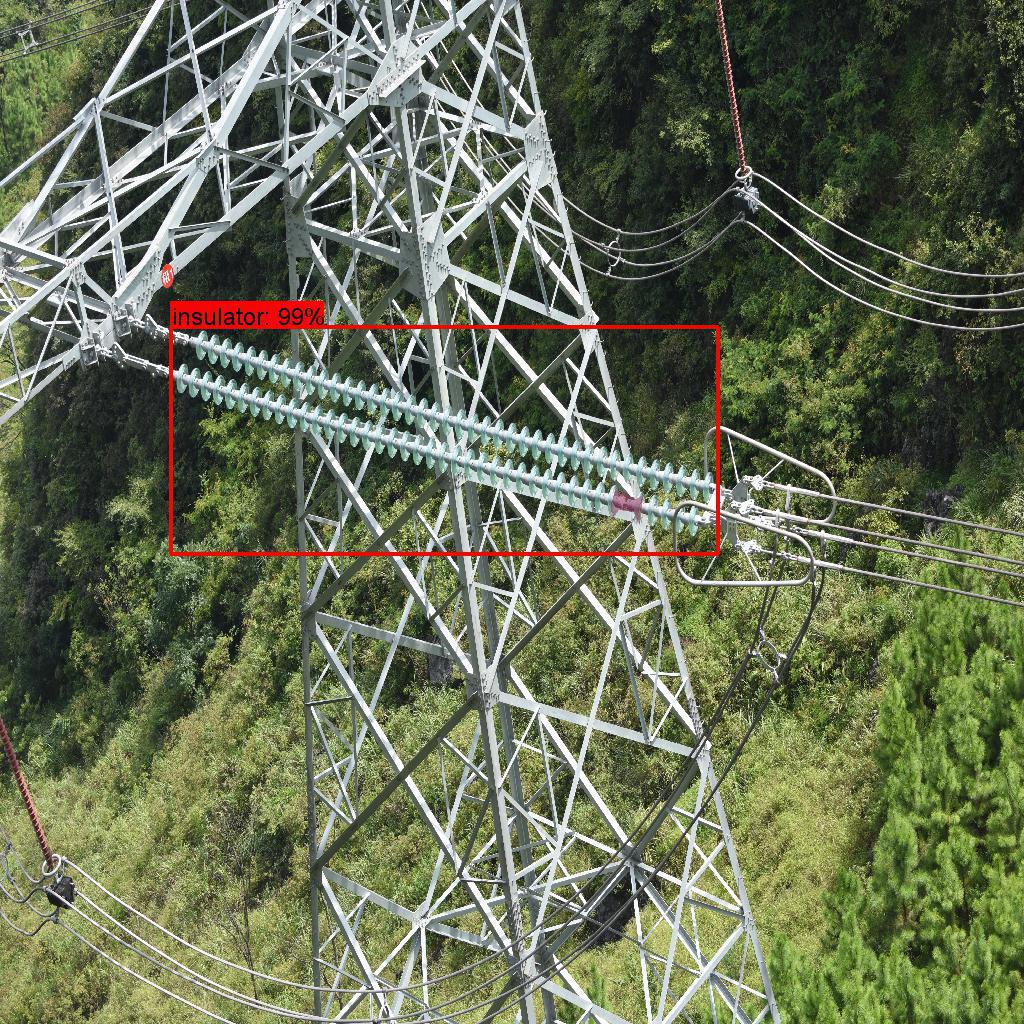}}
 \subfloat{\label{fig9f}\includegraphics[width=0.2\textwidth]{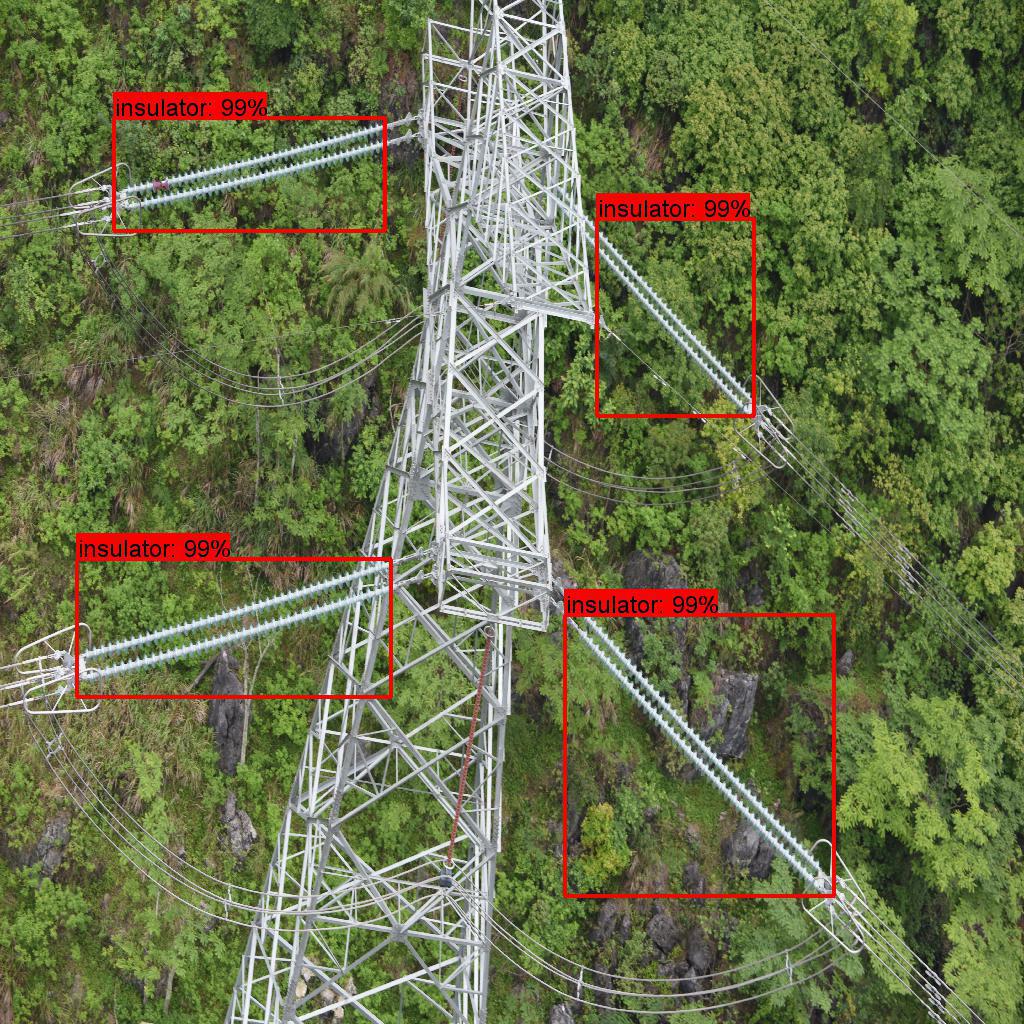}}
 \\
 \subfloat{\label{fig9b}\includegraphics[width=0.2\textwidth]{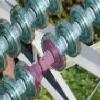}}
 \subfloat{\label{fig9c}\includegraphics[width=0.2\textwidth]{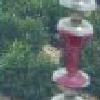}}
 \subfloat{\label{fig9e}\includegraphics[width=0.2\textwidth]{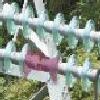}}
 \subfloat{\label{fig9g}\includegraphics[width=0.2\textwidth]{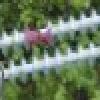}}
 \caption{The self-blast insulator location results. The  magnified images of corresponding self-blast glass insulators are shown in the second line for readers' convenience.}
\label{mask}
 \end{figure*}

Four classification results are shown in Fig.~\ref{mask}. The pixels of failure position is covered  by semitransparent purple masks.  The value of precision is \textbf{0.951} and the value of recall is \textbf{0.955}. The time cost of U-net on a single image is \emph{almost negligible}. This is a fairly ideal result for the broken glass insulator location.

\subsection{Advantage of the Combination of Faster R-CNN and U-net}
To validate the advantage of the combination of these two architectures, the self-blast insulator location  is  carried out with Faster-RCNN or  U-net alone as follows.

The values of the criteria are shown in Tab.~\ref{tab:modelcompare}. U-net alone performs bad, and Faster R-CNN alone performs better but still far away from the proposed method.
\begin{table}
  \centering
  \begin{tabular}{p{3cm}|p{1cm}|p{1cm}}
\hline
  Method &  Precision & Recall \\
  \hline
  U-net & 0.567 & 0.627 \\
  Faster R-CNN & 0.882 & 0.896 \\
  Faster R-CNN plus U-net &\textbf{0.951} & \textbf{0.955} \\
\hline
\end{tabular}
  \caption{The precision and recall of different methods}\label{tab:modelcompare}
\end{table}

\begin{figure*}[htbp]
 \centering
 \subfloat[The result based on U-net alone]{\label{fig11a}
 \includegraphics[width=0.3\textwidth]{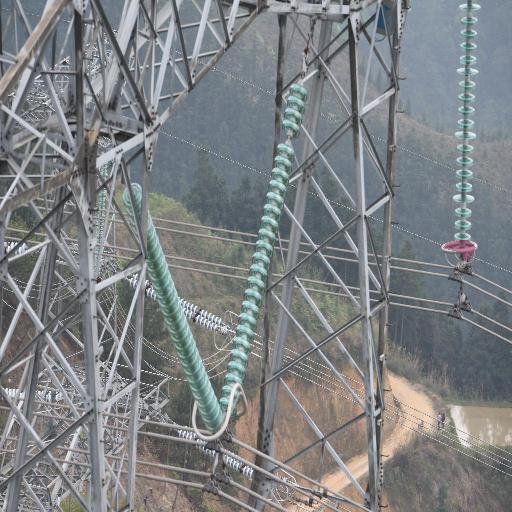}
 }
 \subfloat[The result based on Faster-RCNN alone]{\label{fig11b}
 \includegraphics[width=0.3\textwidth]{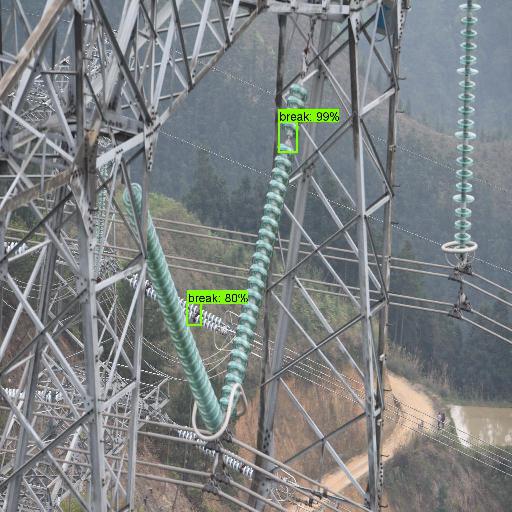}
 }
 \subfloat[The result based on the proposed method]{\label{fig11c}
 \includegraphics[width=0.3\textwidth]{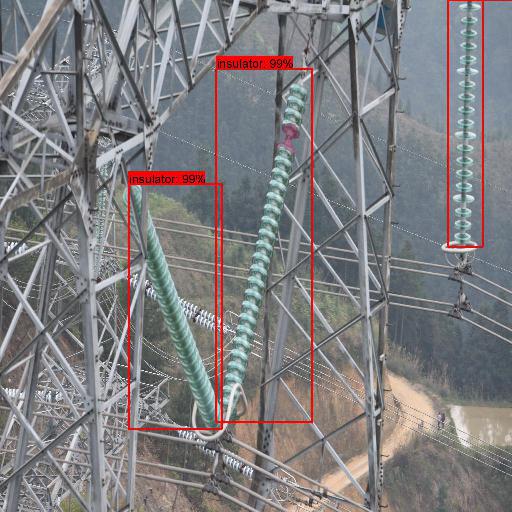}
 }
 \caption{The comparison of three methods}
\label{comparison}
 \end{figure*}

As shown in Fig.\ref{comparison}, an intuitive comparison is  presented among the three methods. Fig.\ref{fig11a} shows that U-net fails to locate the broken insulators. Fig.\ref{fig11b} shows that Faster R-CNN locates the self-blast insulator successfully, however, it seems like that the redundant information interfere with Faster R-CNN and thus an extra false result is obtained. This indicates the importance of the enhancement of SNR. Fig.\ref{fig11c} shows that the proposed  method, which combines Faster R-CNN and U-net, has a good performance.

\subsection{Influence of the Number of Training Samples}
\begin{figure}[htpb]
\centering
\begin{overpic}[scale=0.3]{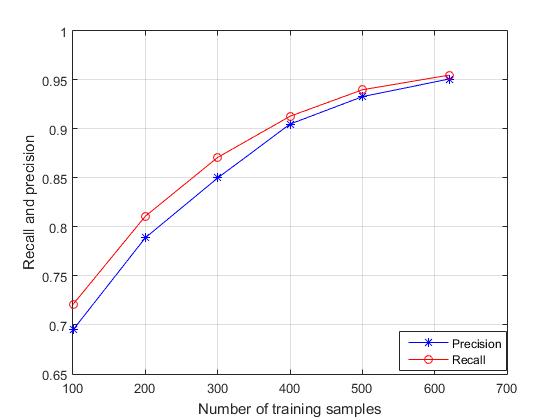}
\end{overpic}
\caption{The criteria with different numbers of training samples}
\label{fig:samplenum}
\end{figure}

In the end of this section, the proposed method is evaluated on the aspect of  training sample  numbers. We set the samples numbers from 100 to 620, and keep the proportion of positive and negative samples about 2:1. The result is shown in Fig.~\ref{fig:samplenum}.

Fig.~\ref{fig:samplenum} shows that the method performs better as the training sample number increase. It means that the method will achieve more accurate results if  more training sample are available.

\section{CONCLUSION}
\label{conclusion}
The methods for aerial images is a bottleneck for broken insulator fault location. This paper proposed a novel self-blast glass insulator location methods based on the deep learning architectures for aerial images. The self-blast glass insulator location problem is formulated into the location and segmentation problem in computer vision. Two \emph{state-of-art} CNN based models, Faster R-CNN and U-net, are introduced. The former is responsible for the location of glass insulator strings so that SNR is enhanced greatly; and the latter enables the precise classification of the pixels in the cropped images of different sizes. The advantages of these two architectures are combined naturally. According to a large number of experimental comparisons,  Faster R-CNN with inception-resnet-v2 pre-trained model is chosen as the optimal object detection architecture so that the best performance of accuracy is obtained while the real-time detection is realized: the value of precision is 0.951, the value of recall is 0.955, and the time cost of a single image is about 800 ms. The experiment results also indicate that the proposed method performs better as the training sample number increases. To reproduce our method, the details of our experiments such as the preprocessing operation, the method of image labelling and the parameter settings is clearly elaborated. Besides, the proposed method can be viewed as a general framework of broken insulator location using aerial images. It is easy to transfer our method to other similar detection problem in this domain.

It is necessary to point out that there is still some room left for our proposed method. In the method, the training and reference of the two networks are dependent---it is not essentially an end-to-end learning method.
Currently, Mask R-CNN~\cite{He2017Mask} is proposed and it combines the detection and segmentation in its architecture by adding FCN with Faster R-CNN. Besides, the results of ~\cite{hu2017learning}  indicates that the pixel label may not be necessary and the coordinates of bounding boxes is enough for Mask R-CNN. In practice, this method will dramatically reduce the cost of time and labor on labelling for pixels. Thus, we will improve our method based on Mask R-CNN in the future work.

\bibliographystyle{IEEEtran}
\bibliography{insulator}

\begin{thebibliography}{10}
\providecommand{\url}[1]{#1}
\csname url@samestyle\endcsname
\providecommand{\newblock}{\relax}
\providecommand{\bibinfo}[2]{#2}
\providecommand{\BIBentrySTDinterwordspacing}{\spaceskip=0pt\relax}
\providecommand{\BIBentryALTinterwordstretchfactor}{4}
\providecommand{\BIBentryALTinterwordspacing}{\spaceskip=\fontdimen2\font plus
\BIBentryALTinterwordstretchfactor\fontdimen3\font minus
  \fontdimen4\font\relax}
\providecommand{\BIBforeignlanguage}[2]{{%
\expandafter\ifx\csname l@#1\endcsname\relax
\typeout{** WARNING: IEEEtran.bst: No hyphenation pattern has been}%
\typeout{** loaded for the language `#1'. Using the pattern for}%
\typeout{** the default language instead.}%
\else
\language=\csname l@#1\endcsname
\fi
#2}}
\providecommand{\BIBdecl}{\relax}
\BIBdecl

\bibitem{Krizhevsky2012ImageNet}
A.~Krizhevsky, I.~Sutskever, and G.~E. Hinton, ``Imagenet classification with
  deep convolutional neural networks,'' \emph{Communications of the Acm},
  vol.~60, no.~2, p. 2012, 2012.

\bibitem{Simonyan2014Very}
K.~Simonyan and A.~Zisserman, ``Very deep convolutional networks for
  large-scale image recognition,'' \emph{Computer Science}, 2014.

\bibitem{Szegedy2014Going}
C.~Szegedy, W.~Liu, Y.~Jia, P.~Sermanet, S.~Reed, D.~Anguelov, D.~Erhan,
  V.~Vanhoucke, and A.~Rabinovich, ``Going deeper with convolutions,'' pp.
  1--9, 2014.

\bibitem{He2015Deep}
K.~He, X.~Zhang, S.~Ren, and J.~Sun, ``Deep residual learning for image
  recognition,'' pp. 770--778, 2015.

\bibitem{Ren2015Faster}
S.~Ren, K.~He, R.~Girshick, and J.~Sun, ``Faster r-cnn: towards real-time
  object detection with region proposal networks,'' in \emph{International
  Conference on Neural Information Processing Systems}, 2015, pp. 91--99.

\bibitem{Shelhamer2017Fully}
E.~Shelhamer, J.~Long, and T.~Darrell, ``Fully convolutional networks for
  semantic segmentation,'' \emph{IEEE Transactions on Pattern Analysis and
  Machine Intelligence}, vol.~39, no.~4, pp. 640--651, 2017.

\bibitem{Zhao2016Multi}
Z.~Zhao, G.~Xu, Y.~Qi, N.~Liu, and T.~Zhang, ``Multi-patch deep features for
  power line insulator status classification from aerial images,'' in
  \emph{International Joint Conference on Neural Networks}, 2016, pp.
  3187--3194.

\bibitem{Liao2017A}
S.~Liao and J.~An, ``A robust insulator detection algorithm based on local
  features and spatial orders for aerial images,'' \emph{IEEE Geoscience and
  Remote Sensing Letters}, vol.~12, no.~5, pp. 963--967, 2017.

\bibitem{Wu2012A}
Q.~Wu, J.~An, and B.~Lin, ``A texture segmentation algorithm based on pca and
  global minimization active contour model for aerial insulator images,''
  \emph{IEEE Journal of Selected Topics in Applied Earth Observations and
  Remote Sensing}, vol.~5, no.~5, pp. 1509--1518, 2012.

\bibitem{Reddy2013Condition}
M.~J.~B. Reddy, C.~B. Karthik, and D.~K. Mohanta, ``Condition monitoring of 11
  kv distribution system insulators incorporating complex imagery using
  combined dost-svm approach,'' \emph{IEEE Transactions on Dielectrics and
  Electrical Insulation}, vol.~20, no.~2, pp. 664--674, 2013.

\bibitem{Ronneberger2015U}
O.~Ronneberger, P.~Fischer, and T.~Brox, ``U-net: Convolutional networks for
  biomedical image segmentation,'' in \emph{International Conference on Medical
  Image Computing and Computer-Assisted Intervention}, 2015, pp. 234--241.

\bibitem{tensorflow2015-whitepaper}
\BIBentryALTinterwordspacing
M.~Abadi, A.~Agarwal, P.~Barham, E.~Brevdo, Z.~Chen, C.~Citro, G.~S. Corrado,
  A.~Davis, J.~Dean, M.~Devin, S.~Ghemawat, I.~Goodfellow, A.~Harp, G.~Irving,
  M.~Isard, Y.~Jia, R.~Jozefowicz, L.~Kaiser, M.~Kudlur, J.~Levenberg,
  D.~Man\'{e}, R.~Monga, S.~Moore, D.~Murray, C.~Olah, M.~Schuster, J.~Shlens,
  B.~Steiner, I.~Sutskever, K.~Talwar, P.~Tucker, V.~Vanhoucke, V.~Vasudevan,
  F.~Vi\'{e}gas, O.~Vinyals, P.~Warden, M.~Wattenberg, M.~Wicke, Y.~Yu, and
  X.~Zheng, ``{TensorFlow}: Large-scale machine learning on heterogeneous
  systems,'' 2015, software available from tensorflow.org. [Online]. Available:
  \url{http://tensorflow.org/}
\BIBentrySTDinterwordspacing

\bibitem{Szegedy2016Inception}
C.~Szegedy, S.~Ioffe, V.~Vanhoucke, and A.~Alemi, ``Inception-v4,
  inception-resnet and the impact of residual connections on learning,'' 2016.

\bibitem{Sutskever2013On}
I.~Sutskever, J.~Martens, G.~Dahl, and G.~Hinton, ``On the importance of
  initialization and momentum in deep learning,'' in \emph{International
  Conference on International Conference on Machine Learning}, 2013, pp.
  III--1139.

\bibitem{Liu2015SSD}
W.~Liu, D.~Anguelov, D.~Erhan, C.~Szegedy, S.~Reed, C.~Y. Fu, and A.~C. Berg,
  ``Ssd: Single shot multibox detector,'' pp. 21--37, 2015.

\bibitem{Dai2016R}
J.~Dai, Y.~Li, K.~He, and J.~Sun, ``R-fcn: Object detection via region-based
  fully convolutional networks,'' 2016.

\bibitem{Szegedy2013Deep}
C.~Szegedy, A.~Toshev, and D.~Erhan, ``Deep neural networks for object
  detection,'' \emph{Advances in Neural Information Processing Systems},
  vol.~26, pp. 2553--2561, 2013.

\bibitem{Sergey2015Batch}
S.~Ioffe and C.~Szegedy, ``Batch normalization: Accelerating deep network
  training by reducing internal covariate shift,'' pp. 448--456, 2015.

\bibitem{Kingma2014Adam}
D.~P. Kingma and J.~Ba, ``Adam: A method for stochastic optimization,''
  \emph{Computer Science}, 2014.

\bibitem{He2017Mask}
K.~He, G.~Gkioxari, P.~Doll¨¢r, and R.~Girshick, ``Mask r-cnn,'' 2017.

\bibitem{hu2017learning}
R.~Hu, P.~Doll{\'a}r, K.~He, T.~Darrell, and R.~Girshick, ``Learning to segment
  every thing,'' \emph{arXiv preprint arXiv:1711.10370}, 2017.

\end{thebibliography}

\end{document}